\pdfoutput=1

\documentclass[11pt]{article}

\usepackage[final]{acl}

\usepackage{times}
\usepackage{latexsym}
\usepackage{graphicx}
\usepackage{amsmath}

\usepackage[T1]{fontenc}

\usepackage[utf8]{inputenc}

\usepackage{microtype}

\usepackage[colorinlistoftodos]{todonotes}
\usepackage[caption=false]{subfig}

\usepackage{listings}
\usepackage{xcolor}
\definecolor{codegreen}{rgb}{0,0.6,0}
\definecolor{codegray}{rgb}{0.5,0.5,0.5}
\definecolor{codepurple}{rgb}{0.58,0,0.82}
\definecolor{backcolour}{rgb}{0.95,0.95,0.92}
\lstdefinestyle{mystyle}{
    backgroundcolor=\color{backcolour},   
    commentstyle=\color{codegreen},
    keywordstyle=\color{magenta},
    numberstyle=\color{codegray},
    stringstyle=\color{codepurple},
    basicstyle=\ttfamily\footnotesize,
    breakatwhitespace=false,         
    breaklines=true,                 
    captionpos=b,                    
    keepspaces=true,                 
    numbers=left,                    
    numbersep=5pt,                  
    showspaces=false,                
    showstringspaces=false,
    showtabs=false,                  
    tabsize=2
}
\lstdefinelanguage{MyDialogue}{
    alsodigit = {:},
    keywords = {Leolani:, Student1:, Student2:, Student3:}
}
\title{Evaluating Agent Interactions Through Episodic Knowledge Graphs}

\author{Selene Báez Santamaría \and Piek Vossen \and Thomas Baier \\
        Vrije University Amsterdam \\ \texttt{\{s.baezsantamaria, p.t.j.m.vossen, t.baier\}@vu.nl}}

\begin{document}
\maketitle
\begin{abstract}
We present a new method based on episodic Knowledge Graphs (eKGs) for evaluating (multimodal) conversational agents in open domains. This graph is generated by interpreting raw signals during conversation and is able to capture the accumulation of knowledge over time. We apply structural and semantic analysis of the resulting graphs and translate the properties into qualitative measures. We compare these measures with existing automatic and manual evaluation metrics commonly used for conversational agents. Our results show that our Knowledge-Graph-based evaluation provides more qualitative insights into interaction and the agent's behavior.
\end{abstract}

\section{Introduction}
In order to develop open-domain conversational agents, it is crucial to have automatic and reproducible ways of evaluating the interaction and the agent's role. However, interaction with people is challenging to evaluate for several reasons: 1) people behave differently in each interaction, 2) people appreciate the interaction for different reasons and aspects, 3) different goals and sub-goals may play a role simultaneously, and 4) personal relationships and past experiences have an impact on every interaction. For these reasons, most evaluations of interactive systems use human judges and questionnaires in analogy to user-satisfaction methods.  

In addition to these questionnaires, conversational systems are often evaluated by comparing system responses to human responses on a turn-by-turn basis, where the prompts and the gold responses are taken from human-human conversations. Standard measures such as BLUE~\cite{papineni2002bleu}, METEOR~\cite{banerjee-lavie-2005-meteor}, BERTscore~\cite{bert-score} test the similarity between the system response and a gold response, whereas USR~\cite{mehri-eskenazi-2020-usr} tests the coherence of the system response to the previous prompt and context. However, these measures do not truly assess the quality of the system's interpretation and the relevance of the prompt, and they \textit{punish systems for being creative and making responses personal}. By detaching prompt-response pairs from the whole conversation, these metrics evaluate the reactivity of an agent, not its ability to engage in a coherent interaction.

\citet{deriu2021survey} mention five general requirements for evaluation: 1) automatic to reduce human labour and subjectivity, 2) repeatable when applied to the same dialogue, 3) correlate with human judgments, 4) differentiated for various strategies, and 5) explainable. None of the existing approaches satisfy all these criteria. In this paper, we demonstrate that graph properties can be used as an additional and independent evaluation of the \textbf{effectiveness} of the communication. This evaluation is an automatic measure of \textbf{semantic quality} that is also explainable and reproducible, meeting three of the five previous requirements.

We present a novel evaluation method that qualifies conversations using an episodic Knowledge Graph (eKG)~\cite{baez-santamaria-etal-2021-emissor}. We calibrate different groups of graph measures in relation to human evaluations and ground truth independent measures. To test our selected metrics, we compare the quality of types of conversations and show that the proposed evaluation framework holds,  regardless of differences in system design. Our contributions are:

\begin{enumerate}
\itemsep0em 
    \item We provide an reference-free and explainable method for evaluating the interaction of conversational agents.
    \item We compare our method to other standard evaluation methods and show its complementary value.
    \item We demonstrate that our method can be applied across multiple conversations and different (types of) participants.
\end{enumerate}


\section{Related work}
\label{sec:related}
Dialogue systems have been studied for several decades and are further developed within Conversational AI systems. In their survey, \citet{deriu2021survey} discuss different types of dialogue systems and how they are evaluated. They conclude that evaluating open conversational agents is an open problem due to the lack of a goal and variable structure. Therefore, evaluation approaches focus on appropriateness and human likeness of responses or specific linguistic properties such as variability, lexical complexity, coherence, correctness, and relevance of system responses. Evaluations can furthermore be done at a turn-level or conversation-level. There is a many-to-many problem in both cases: multiple responses can be correct, multiple dialogues can lead to the right/same result, and every interaction is unique.

Attempts to automate these notions often rely on metrics such as BLUE~\cite{papineni2002bleu}, ROUGE~\cite{lin2004looking}, METEOR~\cite{banerjee-lavie-2005-meteor}, and BERTscore~\cite{bert-score} to measure the similarity of the agent response to one or more ground-truth responses; or they borrow from information retrieval measures when systems need to select the appropriate responses from a set of possible alternatives. In contrast, USR~\cite{mehri-eskenazi-2020-usr} is a new method created by fine-tuning RoBERTa-base~\cite{liu2019roberta} on the training set of Topical-Chat. Whereas METEOR and BERTscore compare a system response to a ground-truth response, USR gives a quality evaluation score without a ground-truth response (reference-free) by measuring the coherence of the system response with the human prompt and the previous context.

Evaluation regimes with ground-truth responses limit agents' "freedom and creativity" to generate other responses that may also fit the purpose. Therefore, it is unsurprising that evaluations often fall back on a posteriori evaluation by human judges. However, human evaluations suffer from several pitfalls: expensive, time-consuming, inconsistent across experiments, difficult to reproduce, and challenging to scale. Researchers tried to harmonize the evaluation criteria to address the inconsistency and lack of coherence in terminology and methodology for human evaluations of open-domain dialogue. \citet{howcroft-etal-2020-twenty} survey of 165 papers with human evaluations reports more than 200 quality criteria (such as Fluency, Accuracy, or Readability.) that have been used in Natural Language Generation. Independently, \citet{fitrianie2019we} analyzed the proceedings of the conference of Intelligent Virtual Agents\footnote{\url{https://dl.acm.org/conference/iva/proceedings}} between 2013 and 2018. They found 189 constructs from 89 questionnaires reported in 81 papers, which they reduced to 19 measurement instruments. Measurements range from how enjoyable, correct and useful to how fluent.

In an attempt to automate evaluations and make them more reproducible and scalable, the \textit{Ninth Dialog System Technology Challenge, Track on Interactive Evaluation of Dialog} (DSTC9, Track 3)\footnote{\url{http://dialog.speech.cs.cmu.edu:8003}} carried out a variety of automatic and human evaluations on 33 systems submissions to the Topical-Chat challenge~\cite{gopalakrishnan2019topical}. Topical-Chat consists of conversations between two Amazon Mechanical Turk workers who were given prior knowledge or information to refer to during their conversation. Systems need to respond to turns from these conversations, replacing one worker. A human evaluation of system responses was done using the questionnaire from the FED dataset~\cite{mehriunsupervised}. An automatic evaluation was done using the measures METEOR, BERTscore, and USR. \citet{gunasekara2020overview} report that the USR (0.3 Spearman) correlates better with human judgments than METEOR (0.23 Spearman) and BERTscore (Spearman 0.22), although they also admit that the correlation is not very high. It is still to be seen how easy USR can be transferred to other dialogues and contexts, as it was trained and tested on Topical-Chat.

We present a reference-free approach that is not based on coherence but measures the interpretability of (multimodal) situations and accumulates these over time. The basic idea is that effective interactions result in rich and high-quality representations that can be measured in a Knowledge Graph. Factors determining the communication effect are the agent's response quality and the collaboration between participants.

\section{Problem formalization}
\label{sec:model}
We represent an interaction as a series of tuples \textit{[t, s, g, f, p]}, such that:

\begin{description}
\small{
\itemsep0em 
    \item[t] \begin{math}\in\end{math} \textbf{T}, a set of time points 
    \item[s] \begin{math}\in\end{math} \textbf{S}, a set of situations
    \item[g] \begin{math}\in\end{math} \textbf{G}, a set of graphs
    \item[f] \begin{math}\in\end{math} \textbf{F}, a set of unknown features part of situation \textbf{s}
    \item[p] \begin{math}\in\end{math} \textbf{P}, a set of defined properties that can make up graph \textbf{g}  
}
\end{description}

A graph \textbf{g} represents the interpretation of a sequence of situations \textbf{s} at time point \textbf{t}. Each situation \textbf{s} can be modeled as a bundle of unknown features \textbf{f} and each graph \textbf{g} as a set of properties \textbf{p} that are defined a priori. To quality the conversations, we measure how many and which properties \textbf{p} are extracted from each turn and the cumulative effect of adding these properties over time to the graph \textbf{g}. The instrument's effectiveness in measuring quality depends on the ability to detect the properties \textbf{p} given the features \textbf{f}. The effectiveness of the communication depends on the predefined properties \textbf{p} that are chosen for the evaluation. 

The properties used to define the quality of the conversation can be 1) mathematical, e.g. measuring the average degree and sparseness, 2) semantic, e.g. number and type of triples, 3) knowledge integrity, e.g. conflicts, outliers, analogies, completeness, 4) subjective values, e.g. sentiment and emotion, certainty, trust, and 5) dialogue properties, e.g. turn-property ratio's, utterance type distribution and density, style and quality of expression. These measures depend on the capability to extract \textbf{p} from the unknown features \textbf{f} -implicit in image, audio and text signals- and the modelling of these properties in the knowledge graph \textbf{g}. An interaction will result in a series of graphs over time. A cumulative graph can be seen as an episodic Knowledge Graph (eKG) \cite{baez-santamaria-etal-2021-emissor} for which the qualitative evaluation over time can provide valuable additional insight.

In this paper, we implement the above formal model and propose a set of properties \textbf{p}, defined as RDF triples, that correlate with human judgements and can be used across different conversational setups and for different property detection systems.

\subsection{Model}
\begin{figure}[!ht]
    \centering
    \includegraphics[trim={0 0 0 0},clip,width=.45\textwidth]{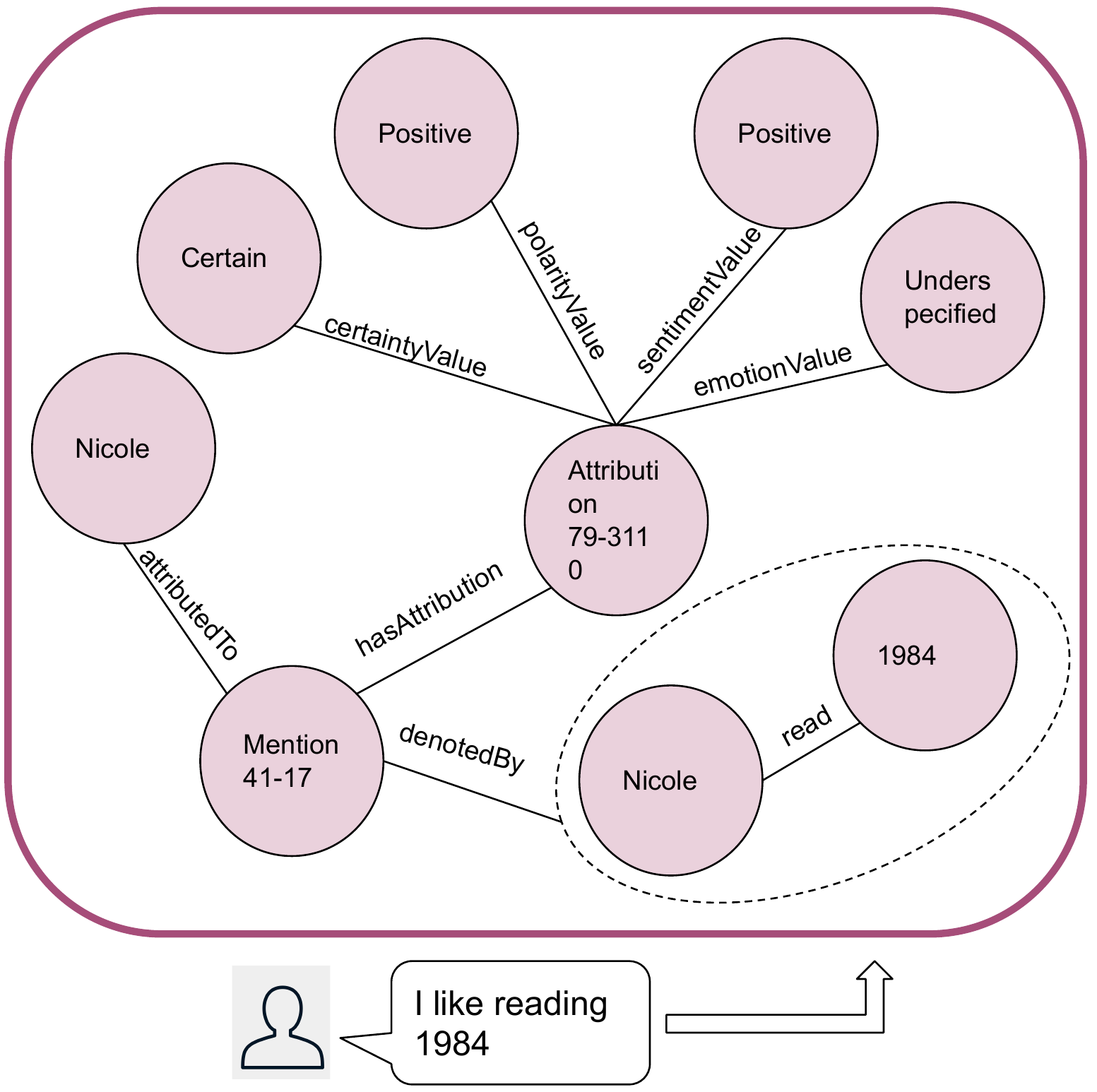}
    \caption{Example of how an utterance is converted into an episodic graph with source perspective values.}
    \label{fig:utterance-graph}
\end{figure}

Figure \ref{fig:utterance-graph} shows the interpretation of the statement "I like reading 1984" in an eKG. The core triple reflects that "Nicole" "reads" something labelled as "1984". The triple itself is a named graph representing a claim. The claim is mentioned (denotedBy) by the speaker Nicole (attributedTo) and perspective values are attributed to this mention, such as sentiment:positive, polarity:positive, certainty:certain. The model can represent multiple mentions of the same triple, with different perspective values and/or attributed to different speakers. 

\begin{figure*}[!ht]
    \centering
    \includegraphics[trim={0 0 0 0},clip,width=.95\textwidth]{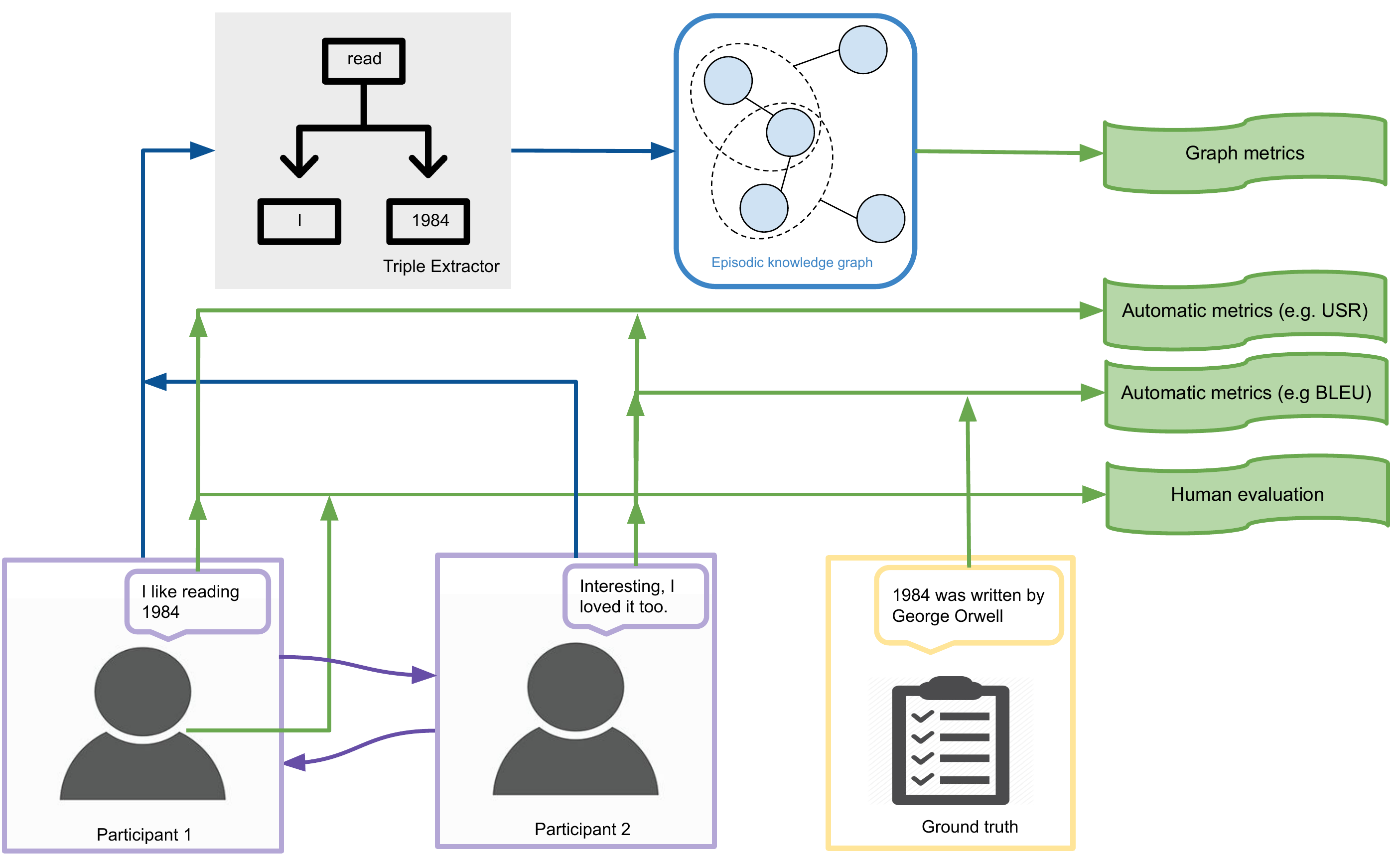}
    \caption{Schematic representation of interaction process (in purple), golden data (in yellow), EKG construction (in blue) and evaluation (in green). Arrows flowing into the metrics determine the components needed for their calculation. }
    \label{fig:evaluation-schema}
\end{figure*}

\section{Capturing interaction and episodic knowledge}
\label{sec:method}
In order to apply graph metrics, we need to generate properties from the (multimodal) signals produced during the interaction. As we represent these properties as RDF triples, we rely on a text-to-triple instrument to detect these properties. This instrument's effectiveness determines the evaluation's depth, precision and richness; therefore, the most precise and standardised instrument is preferable. However, using the same instrument across interactions, we can compare the interactions and draw conclusions about its conversational variables. 

Another factor for our evaluation is the graph properties modelled in the eKG. Some properties are generic and can be measured in any graph, whereas other properties depend on the semantics of the data model. In the following subsections, we discuss conversational variables when comparing interactions, the metrics that can be applied to the generated graphs and how they depend on specific property types.

\subsection{Conversational variables}
\label{sec:variables}
Our framework is agnostic to an interaction setup. We can thus have various combinations of prompt and response participants, and we can use different triples extractors, as shown in Figure~\ref{fig:evaluation-schema}. Schematically, the interlocutors (P1 and P2) both produce conversational signals to which we can apply triple-extraction, updating the eKG. Our evaluation framework is applied to this eKG, which is reference-free and considers the interaction between both interlocutors. This differs from other evaluation frameworks, such as USR, which only evaluates P2 as a coherent response to P1, and BLUE, METEOR and BERTscore, which evaluate P2 against a ground-truth response. 

\paragraph{Triple extractor}
Regardless of the interlocutors, we can apply triple extraction to the utterances a-posteriori and derive an eKG from the communication. The extracted triples represent factoid information and possibly the speakers' perspectives. Some approaches to extract triples are: StandforOpenIE~\cite{angeli2015leveraging}, spaCy's Dependency parser~\cite{vasiliev2020natural}, or tailored Context Free Grammars (CFG).

\paragraph{Agents}
As for the type of participants, we can take recorded human-human dialogues and apply triple extraction to each prompt from an actor. Similarly, machine-machine conversations can be generated where a chatbot mimics human input as a \textit{prompt agent} and another chatbot functions as the \textit{response agent}.

\begin{table*}[!hbt]
    \centering
    \scriptsize{
    \caption{Statistics for 9 conversations with the number of turns, claims and perspectives. The conversation effect is measured by the average claim-triples (claim density) and perspective-triples (perspective density) per turn.}
    \label{table:conversation overview}
    \begin{tabular}{|l|c|c|c|c|c|c|c|c|}
    \hline
     \textbf{Category}  &  \textbf{P1}  &  \textbf{P2}  &  \textbf{Triple Extractor}  &  \textbf{Turns}  &  \textbf{Claims}  &  \textbf{Claim density}  &  \textbf{Perspective}  &  \textbf{Perspective density} \\ \hline \hline
     Human - Machine & Student & Leolani & CFG & 83 & 27 & 0.33 & 23 & 0.28  \\ \hline
     Human - Machine & Student & Leolani & CFG & 57 & 18 & 0.32 & 14 & 0.25 \\ \hline
     Human - Machine & Student & Leolani & CFG & 45 & 17 & 0.38 & 14 & 0.31  \\ \hline
     Human - Machine & Student & Leolani & CFG & 55 & 14 & 0.25 & 11 & 0.20  \\ \hline
     Human - Machine & Student & Leolani & CFG & 78 & 21 & 0.27 & 16 & 0.21  \\ \hline
     Human - Machine & Student & Leolani & CFG & 80 & 22 & 0.28 & 17 & 0.21  \\ \hline
    \hline
     Machine - Machine & Blenderbot & Leolani & CFG-spacy & 298 & 6 & 0.02 & 4 & 0.01 \\ \hline
     Machine - Machine & Blenderbot & Eliza & CFG-spacy & 207 & 24 & 0.12 & 22 & 0.11 \\ \hline
    \hline
     Human - Human & Monica & Chandler & Stanford-OIE & 243 & 109 & 0.44 & 0 & 0.00 \\ \hline
    \end{tabular}}
\end{table*}

\subsection{Graph metrics}
\label{sec:graph-metrics}
Our formal model allows us to evaluate conversations under different frameworks: as a mathematical object (group A), as an RDF knowledge representation tool (group B), and as an eKG hosting the accumulation of interactions (group C). For groups A and B we used an implementation by \citet{pernisch2020chimp} while for groups C we implemented the metrics using \textit{rdflib}. 

After an exploratory analysis with 62 graph metrics (A-15, B-27, C-20), we select a subset of 24, as many of these metrics are compositional and therefore correlate highly and mostly reflect the same insights. Our selected Group A metrics include volume (number of nodes and edges), centrality (average node degree, degree centrality, and closeness), connectivity (average degree connectivity and assortativity), clique (number of strong connected components), entropy (centrality and closeness entropy) and density (sparseness). Group B metrics include volume (number of axioms) and density (average population). Finally, Group C metrics include density (ratios comparing claims to triples, perspectives to triples, conflicts to triples, perspectives to claims, mentions to claims, conflicts to claims), and interaction (average perspectives per claim, mentions per claim, turns per interaction, claims per source and perspectives per source). 

The above measures can be applied to the evolving eKG during the interaction or a posteriori. Although these measures were not originally intended as quality measures for interaction, we hypothesize that some of these measures can be used to characterize a conversation, and the resulting graphs can be compared independently. For example, the average claims per source may signal how much information the agent is getting per person it interacts with, the ratio of mentions to claims signals how much a factoid has been repeated in conversation, while the ratio of perspectives to claims signals how much diversity of opinions or sentiment has been expressed on claimed factoids. 

\section{Experimental setup}
\label{sec:exp-setup}

Table \ref{table:conversation overview} shows an overview of the conversations analyzed. Details on the artificial agents and triple extractors can be found in the Appendix as well as an in-depth analysis of the triples extracted. We calculate the number of claims and the number of perspectives from each session. To measure the overall effectiveness in interpreting the conversation, we derive the density of claims and perspectives per turn\footnote{Note that the number of turns includes Participant 2's responses, while the extraction focuses on claims and perspectives from P1. About half of the total number of turns are utterances from the students, which makes densities of 0.25 and higher still effective.}, which can be seen as the first crude measure of quality.

\paragraph{Human - Machine conversations} Three groups of students had two conversations with Leolani~\cite{vossen2018leolani} during which they had to introduce themselves. Students were instructed to converse for 5 to 10 minutes or 30-40 turns per conversation. They were also instructed about the type of sentences and expressions from which the agent could extract triples to make the conversations more successful. Conversations will be more cumbersome in real open settings where users do not know what the agent can understand. 
\paragraph{Machine - Machine conversations} We set up dialogues between Blenderbot~\cite{roller2020recipes} and Leolani, and Blenderbot with Eliza~\cite{weizenbaum1966eliza}, where we extract triples using a tailored Context Free Grammar (CFG) and spaCy's dependency parses.

\paragraph{Human - Human conversations} We take all dyadic dialogues in the Friends dataset~\cite{poria2018meld} between Monica and Chandler and extract triples using StanfordOIE~\cite{angeli2015leveraging}.

\section{Comparing within conversational variables}
\label{sec:exp1}
To measure the quality of multiple interactions across different participants with the same agent, we collected human judgements of the human-machine conversations and compared these with the graph metrics groups. 

\subsection{Human evaluations}
\label{sec:human-results}
The students evaluated the agent's responses in their conversation using the DSTC9 Track3 challenge evaluation metrics \cite{mehri2022interactive}, which form three submetric groups: 1) enjoyability (Interesting, Engaging, Specific, Relevant), 2) semantic correctness (Correct, Semantically Appropriate, Understandable), and 3) fluency. Each conversation got between two to four evaluations where students score each turn for all submetrics and overall score. Table \ref{table:human_eval} shows the aggregated results, showing the averaged scores for all six conversations. Although Leolani has limited communication skills, most ratings fall above mid-range. 

\begin{table}[!ht]
    \centering
    \scriptsize{
    \caption{Human ratings of six conversations.  Score range: 1 (very bad) to 5 (very good).}
     \label{table:human_eval}
    \begin{tabular}{|l|c|c|c|c|c|c|}
    \hline
          ~ & \multicolumn{6}{|c|}{\textbf{Conversations}}  \\ \hline

        ~ & \textbf{1.1} & \textbf{1.2} & \textbf{2.1} & \textbf{2.2} & \textbf{4.1} & \textbf{4.2} \\ \hline
        Interesting & 2.65 & 2.97 & 1.56 & 1.60 & \textbf{2.98} & 1.91 \\ \hline
        Engaging & 2.79 & \textbf{3.12} & 1.92 & 2.69 & 2.94 & 2.22 \\ \hline
        Specific & \textbf{2.94} & 2.84 & 2.02 & 2.31 & 2.68 & 2.02\\ \hline
        Relevant & 3.25 & \textbf{3.86 }& 2.83 & 2.76 & 2.97 & 2.25 \\ \hline   \hline
        Correct & 3.15 & \textbf{3.93} & 2.63 & 2.55 & 3.01 & 2.20 \\ \hline 
        Semantic appr. & 3.06 & \textbf{3.91} & 2.44 & 2.48 & 2.98 & 2.27 \\ \hline
        Understandable & 3.75 & \textbf{4.06} & 3.33 & 3.05 & 3.31 & 2.97 \\ \hline  \hline
        Fluent & 3.64 & \textbf{4.13} & 2.67 & 2.45 & 3.16 & 3.08 \\ \hline  \hline
        Average submetrics & 3.15 & \textbf{3.60} & 2.42 & 2.49 & 3.00 & 2.37 \\ \hline  \hline
        Overall hum. & 3.16 & \textbf{3.41} & 2.35 & 2.57 & 2.74 & 2.12 \\ \hline
    \end{tabular}}
\end{table}

Table \ref{table:human_MSE} shows the average overall score (2.73) and the average over the submetrics (2.84), which hints that the submetrics comprehensively indicate the overall appreciation. The submetrics vary across but are close to the overall average. The enjoyable submetrics score lower than the semantic correctness and fluency ones.


\subsection{Automatic evaluations}
\label{sec:automatic-results}
Following DSTC9-track3, we scored the agent responses using the USR model ("adamlin/usr-topicalchat-roberta\_ft"). We implemented a likelihood score (USR LLH) that averages the masked-task prediction of the model for every token in the agent response given the preceding utterances as context (up to 300 characters). For every token, we get the top 20 predictions to get the token's score, or a score of zero if it was not listed. We also average the likelihood of the highest-scoring token (USR MAX) as the perfect response according to the pre-trained model. We normalized the USR scores to a 5-point Likert scale to match it with the human ratings and averaged over all responses and conversations. The response of our agent scores significant lower (USR LLH=1.68) than the Overall Human Rating of the conversations (2.73). The maximal possible score by USR is close the to Overall Human Rating (USR MAX=2.78).

\begin{table}[!ht]
    \centering
    \scriptsize{
    \caption{Average human ratings.  Score range: 1 (very bad) to 5 (very good). Mean Squared Error (MSE) for submetric and USR metric averages against the overall human rating.}
     \label{table:human_MSE}
    \begin{tabular}{|l|c||c|c|c|c|}
    \hline
          ~ &  \textbf{} &  \multicolumn{3}{|c|}{\textbf{MSE against overall human score}} \\ \hline

        ~ & \textbf{Avg} & \textbf{Human} & \textbf{USR LLH} & \textbf{USR MAX} \\ \hline
        Interesting & 2.28 & 0.05 & 0.15 & 0.16 \\ \hline
        Engaging & 2.61 & 0.06 & 0.13 & 0.19 \\ \hline
        Specific & 2.47 & 0.06 & 0.13 & 0.16 \\ \hline
        Relevant & 2.99 & 0.04 & 0.13 & 0.16 \\ \hline   \hline
        Correct & 2.91 & 0.03 & 0.14 & 0.13 \\ \hline 
        Semantic appr. & 2.86 & 0.03 & 0.14 & 0.15 \\ \hline
        Understandable & 3.41 & 0.08 & 0.14 & 0.15 \\ \hline  \hline
        Fluent & 3.19 & 0.09 & 0.19 & 0.18 \\ \hline  \hline
        Average submetrics & 2.84 & 0.05 & 0.19 & 0.18 \\ \hline  \hline
        Overall hum. & 2.73 & ~ & 0.15 & 0.16 \\ \hline
    \end{tabular}}
\end{table}

We measured the Mean Squared Error (MSE) by comparing the USR scores and the human submetrics against the Overall Human Rating. The MSE scores for the human submetrics below 1 point. The USR LLH and USR MAX scores are 2 to 3 times higher but remain below 2 points, which means they deviate less than average from the human norm. Finally, our agent response (USR LLH) in most cases correlates better than the most likely predicted tokens from the model itself (USR MAX). 

\subsection{Episodic knowledge graph evaluations}
\label{sec:graph-results}
The previous evaluations (human questionnaires and automatic USR) do not evaluate the quality of the knowledge communicated. To that end, we apply the graph measures described in Section \ref{sec:graph-metrics} to the eKGs of the student conversations. We compute the correlations between the graph metrics and the human and automatic metrics for each student conversation (Appendix Figure \ref{fig:correl}). Two patterns are visible: metric group A correlates more strongly with human evaluations, while metric group B correlates with automatic evaluation. Seven of the human metrics correlate the most with the average degree per node in the eKG. The other two human evaluations, Overall Human Rating and Relevance, correlate the most to sparseness.

\section{Comparing across conversational variables}
\label{sec:exp2}

Three major factors determine the resulting graph: 1) the (human) participant, 2) the agent's capability to understand the prompts, and 3) the agent's capability to respond adequately. Our eKG-based evaluation, therefore, genuinely evaluates interaction from both ends. This makes it possible to evaluate interaction across different (types of) people with the same and/or different agents, resulting in different graphs due to the human input and/or the agents' capabilities. 

\subsection{Correlation with human judgements}
\label{sec:overlap-with-human}

Table \ref{table:graph_eval} shows the values between human evaluations, automatic evaluations, and the graph metric with highest correlation.\footnote{Chatbot conversations have not been evaluated manually.} For 6/9 human metrics, including Overall Human Rating, two graph metrics correlate more strongly than the USR metrics. We interpret this as evidence for these two graph evaluations to approximate human evaluations instead of USR evaluations. Recall that \citet{gunasekara2020overview} reported a correlation of 0.3 for USR concerning the Topical-Chat evaluation, which is higher than the score obtained for our conversations. Since USR was fine-tuned on Topical-Chat training data, it is expected to reflect stronger coherence relations on these conversations. Furthermore, Topical-Chat consists of human-human conversations replaced by system responses, whereas our conversations are naturally-born agent conversations that partly come from the inner drives of the agent. The task to generate an appropriate response is more challenging for our agent compared to fine-tuned language models that mimic human responses.

\begin{table}[!ht]
    \centering
    \scriptsize{
    \caption{Human evaluation, automatic scores and the most correlated graph metric.}
    \label{table:graph_eval}
    \begin{tabular}{|l|l|l|l|}
    \hline
     & \multicolumn{1}{|p{1cm}|}{Average degree} & \multicolumn{1}{|p{1cm}|}{Sparseness} & \multicolumn{1}{|p{1cm}|}{USR LLH}  \\ \hline \hline
    Interesting & 0.088 & 0.077 & \textbf{0.148}  \\ \hline
    Engaging & \textbf{0.158} & 0.145 & 0.076  \\ \hline
    Specific & \textbf{0.124} & 0.067 & 0.072 \\ \hline
    Relevant & 0.055 & 0.062 & \textbf{0.091}  \\ \hline
    \hline
    Correct & 0.071 & 0.040 & \textbf{0.128}  \\ \hline
    Semantically Appropriate & \textbf{0.124} & 0.076 & 0.053  \\ \hline
    Understandable & \textbf{0.119} & 0.050 & -0.013  \\ \hline
    \hline
    Fluent & \textbf{0.184} & -0.061 & -0.039  \\ \hline
    \hline
    Overall Human Rating & 0.120 & \textbf{0.194} & 0.088  \\ \hline 
    \end{tabular}}
\end{table}

\begin{figure*}[!ht]
\centering
    \subfloat[Average node degree.
        \label{fig:degree}]{
        \includegraphics[trim={0 0 0 0.5cm},clip,width=0.5\textwidth]{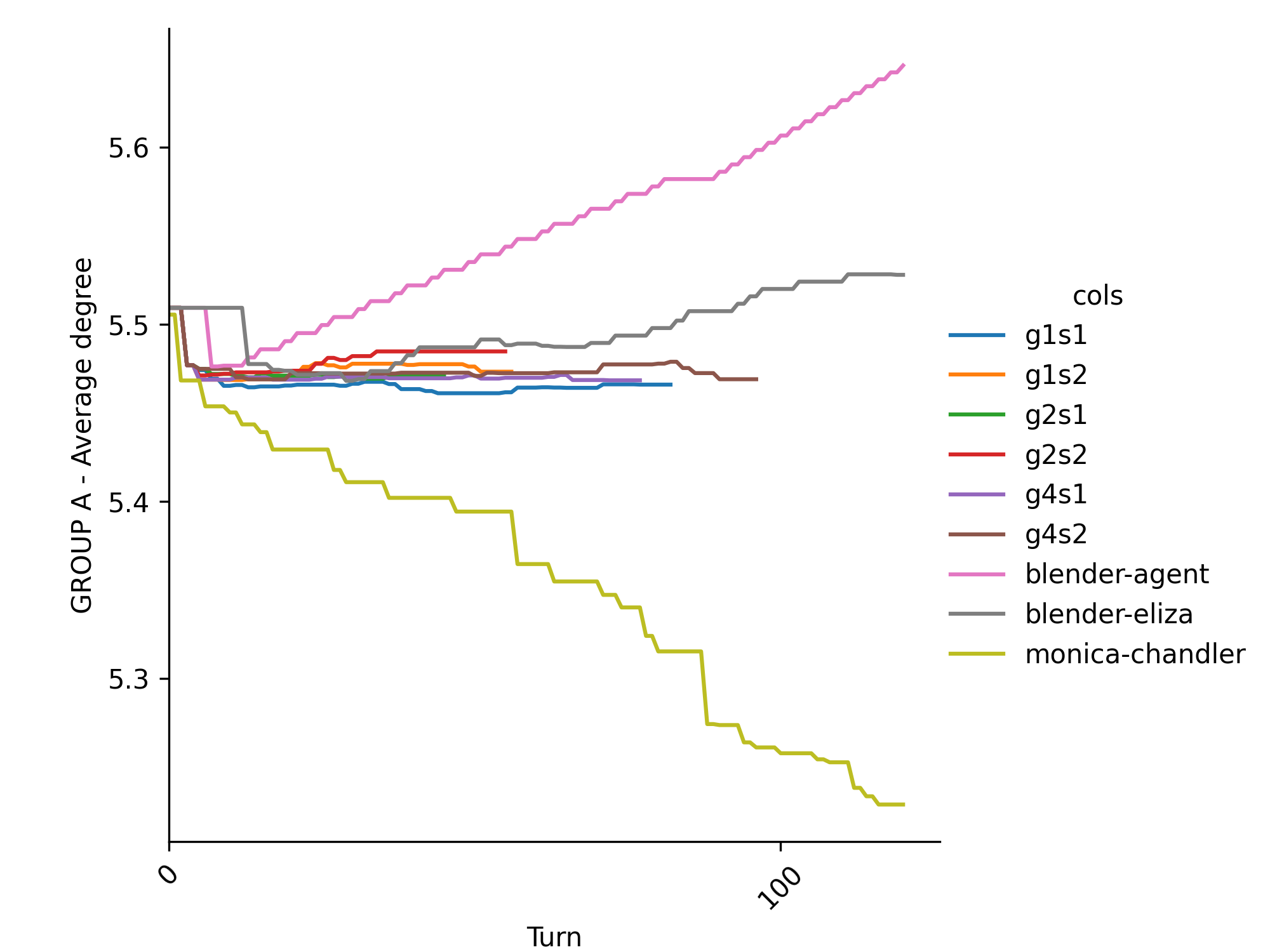}%
    } ~
    \subfloat[Sparseness.
        \label{fig:sparseness}]{
        \includegraphics[trim={0 0 0 .5cm},clip,width=0.5\textwidth]{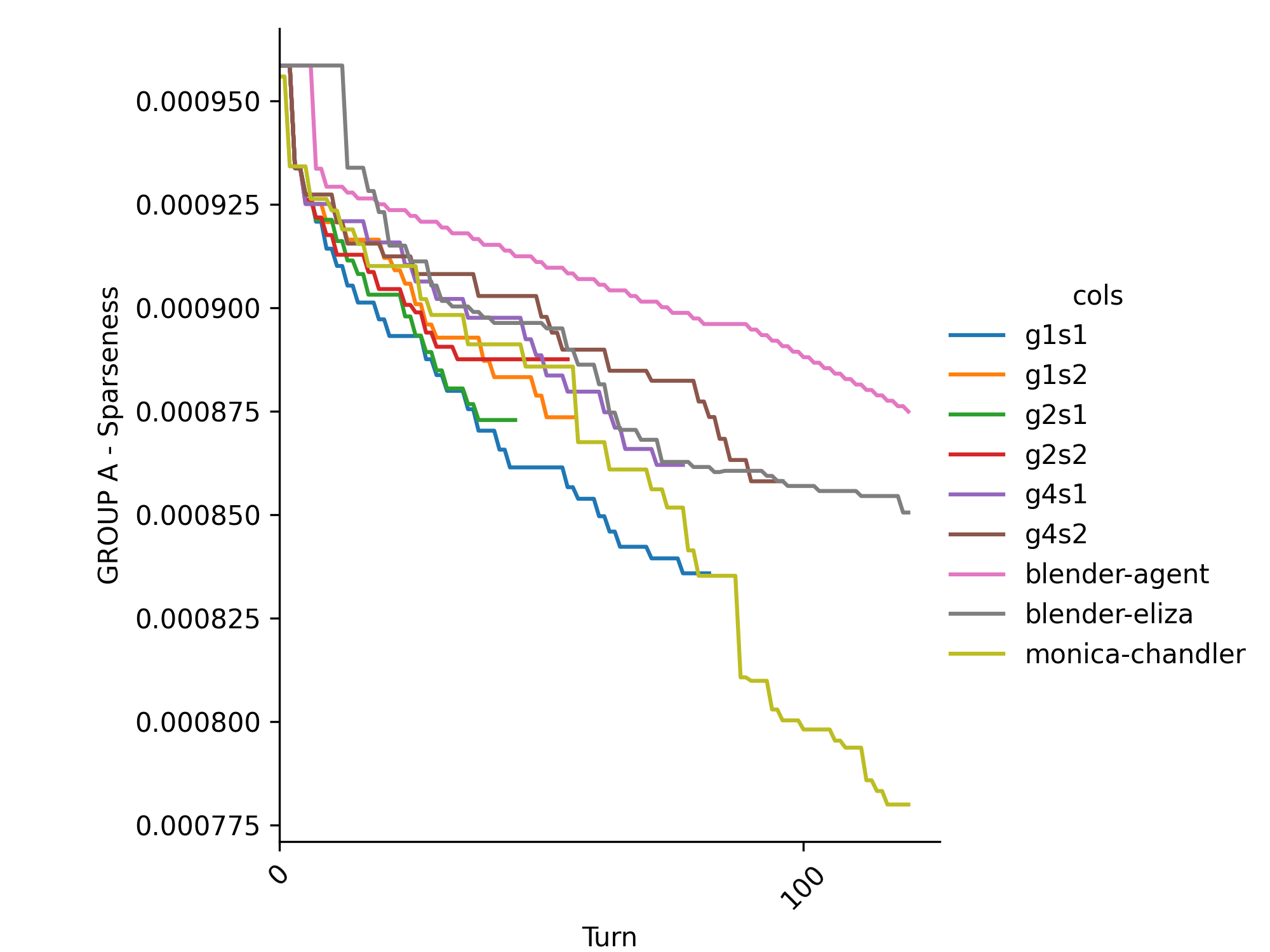}%
    }
    \\
        \subfloat[Ratio of mentions to claims.
        \label{fig:mention-to-claim}]{
        \includegraphics[trim={0 0 0 0.5cm},clip,width=0.5\textwidth]{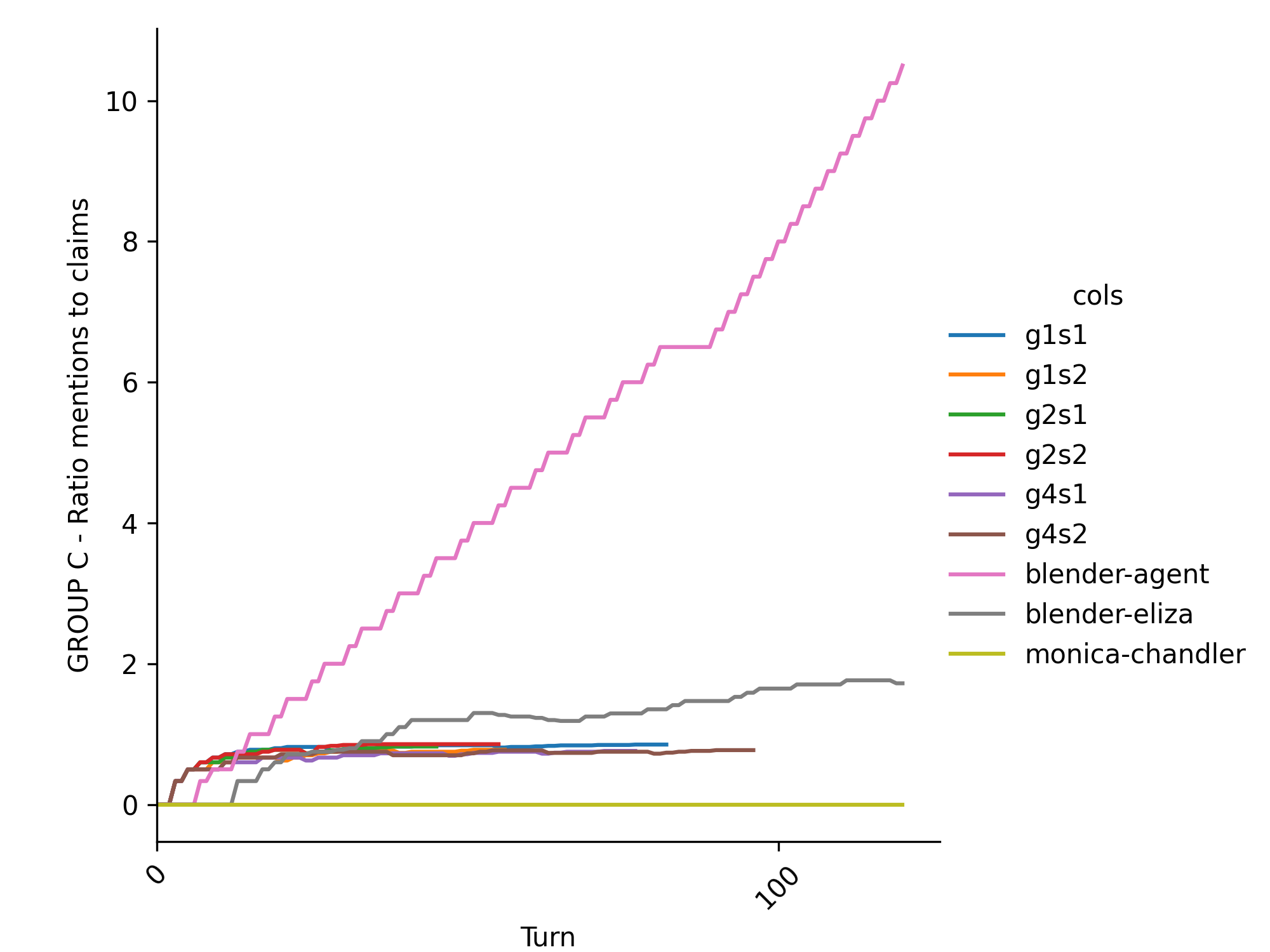}%
    } ~
    \subfloat[Ratio of perspectives to claims.
        \label{fig:perspective-to-claim}]{
        \includegraphics[trim={0 0 0 .5cm},clip,width=0.5\textwidth]{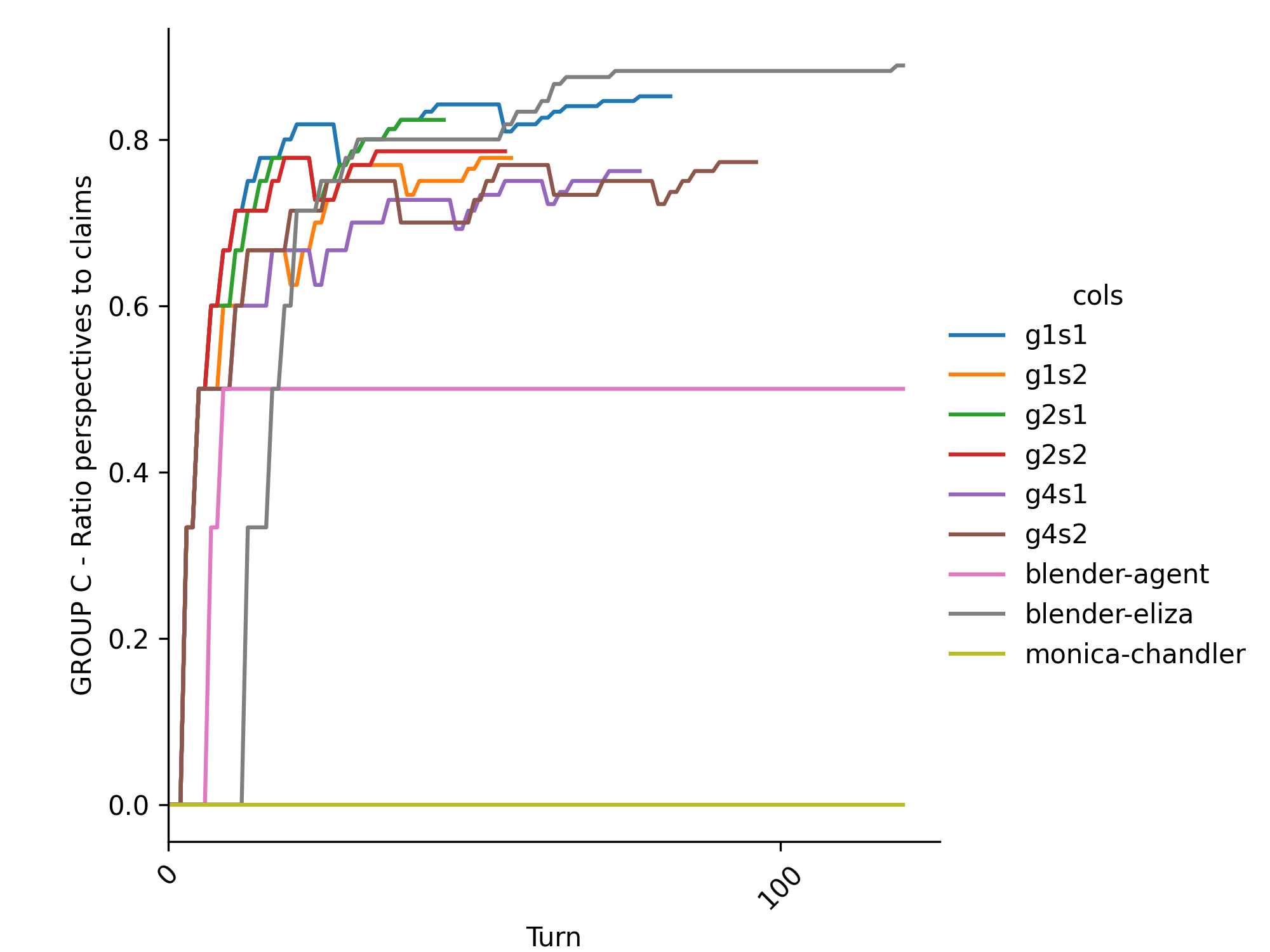}
    }
\caption{Selected graph metrics progression over turns for all student conversations}\label{fig:graph-progression}
\end{figure*}

\paragraph{Average degree}
Average node degree reflects how many edges a node on the eKG has. Figure~\ref{fig:degree}  exposes a relation between average degree and fluency. Conversations 1.2 and 1.1 scored the highest on fluency, while conversation 1.1 shows the lowest average degree curve. In contrast, conversation 2.2 scored the lowest on fluency and showed an incremental behaviour for this metric. Both Blenderbot conversations show a steep increase of the average degree as the conversation proceeds. Manual inspection of these dialogues reveals that Blenderbot becomes repetitive after several turns, resulting in an extreme increase of the average degree. Responses from Leolani are more repetitive than responses from Eliza, which is consistent as Blenderbot is trained with PersonaChat~\cite{zhang2019dialogpt}, Empathetic Dialogues~\cite{rashkin2018towards} and Wikipedia topic conversations~\cite{dinan2018wizard}, making Blenderbot responsive for Eliza's empathic prompts to talk about personal relations and emotions. The responses from Leolani, on the other hand, are based on its drives which can be more obscure and less "human", causing Blenderbot to fall back on standard responses rapidly. Finally, the conversation between Monica and Chandler shows a different pattern, where the degree drops linearly. The fast decreasing curve can be explained by high fluency between (scripted) human-human dialogues.


\paragraph{Sparseness}
Sparseness reflects how well connected a graph is. Figure \ref{fig:sparseness} shows that, for all conversations, sparseness decreases as the conversation proceeds and the Overall Human Rating get higher as the eKG gets less sparse. Conversation 4.2 has the slowest decaying curve, while conversation 1.1 decays the fastest. Conversation 2.2 plateaus for a few turns, which is reflected by having the lowest Overall Human Rating. Since all eKGs have the same sparseness starting point, this suggests that conversations that fail to expand these initially dense graphs might not be successful. The Blenderbot-Leolani conversation is less steep, suggesting that it is less successful. The Blenderbot-Eliza conversation appears to be very similar in decreasing sparseness to the student-Leolani conversations, thus confirming that Blenderbot and Eliza are well aligned. The most effective conversation is shown by Monica and Chandler, having a curve that decays to the same level as conversation 1.1, but decays further.

\subsection{Complementing human judgements}
\label{sec:complement-to-human}
While some metrics correlate with human judgments, other graph metrics complement human evaluations. Evaluating conversations as eKGs allows to observe how much knowledge has been accumulated so far, how much diversity of opinions and conflicting information has been encountered, how often the same factoids are mentioned, how long conversations are, and how much knowledge has been acquired per source. 

\paragraph{Ratio of mentions to claims}
Figure \ref{fig:mention-to-claim} shows the ratio of mentions to claims, which relates to how often the same topics are discussed in conversation. An increasing curve implies a repetitive conversation. The blender-Leolani conversation has the steepest curve, as the conversation stagnated with BlenderBot repeating the same factoid ("I have a dog"). In contrast, the Blenderbot-Eliza conversation goes well, almost like the student-Leolani conversations but still repetitive. Once again, the highest quality conversation is the human-human, with the lowest ratio.

\paragraph{Ratio of perspectives to claims}
Figure \ref{fig:perspective-to-claim} shows the ratio of perspectives to claims, where higher means conversations contain more diverse views on the same topics. On the contrary, a lower ratio represents a series of broad conversations on their topics limited to the views of a few, if not only a single source. The lowest curve belongs to the Chandler-Monica conversation; due to using StanfordOIE, which does not extract perspectives.

\section{Conclusion}
\label{sec:conclusion}
We presented a new method for evaluating dyadic interactions that does not require a ground-truth interaction or a human judgment a posteriori. Our method analyses the episodic Knowledge Graph that results from interpreting prompts. Like the USR score, our method is automatic and reference-free. However, we provided evidence that our evaluation correlates better with human judgments and gives deeper insight into the knowledge built up from the conversation. We also conclude that the graph metrics provide nuanced information about the growth of knowledge resulting from the interaction. Although we cannot yet say anything about the absolute score of an interaction, we can compare different interactions based on the resulting graph and observe differences in graph properties that may or may not be desirable. 

Note that this line of work aims to approximate human evaluations in a cost-free manner. Yet, human evaluations are generally highly subjective and not reproducible to evaluate a conversation. Thus, even though Overall Human Ratings highly correlate to eKG sparseness, outliers arise due to the impact of individual judgments per student group.

In future work, we want to validate these metrics by evaluating benchmark datasets, similar to the methodology by \citet{li2019acute}. We also want to direct the conversations to aim for certain types of knowledge and perspectives to validate that our metrics can detect such different intentions. We want to demonstrate that we can associate graph structures with functionality, e.g., having broad or deep knowledge of subjects, being able to direct people to trustworthy informants, or directing people to shared interests.

\paragraph*{Supplemental Material Statement:}Code and data are publicly available from
\url{https://github.com/selBaez/evaluating-conversations-as-ekg.git}. 

%
\paragraph*{Acknowledgements}
This research was funded by the Vrije Universiteit Amsterdam and the Netherlands Organisation for Scientific Research (NWO) via the \textit{Spinoza} grant (SPI 63-260) awarded to Piek Vossen,  the \textit{Hybrid Intelligence Centre} via the Zwaartekracht grant (024.004.022) and the research project (9406.D1.19.005) \textit{Communicating with and Relating to Social Robots: Alice Meets Leolani}.

\bibliography{main,anthology,custom}

\begin{thebibliography}{26}
\expandafter\ifx\csname natexlab\endcsname\relax\def\natexlab#1{#1}\fi

\bibitem[{Angeli et~al.(2015)Angeli, Premkumar, and
  Manning}]{angeli2015leveraging}
Gabor Angeli, Melvin Jose~Johnson Premkumar, and Christopher~D Manning. 2015.
\newblock Leveraging linguistic structure for open domain information
  extraction.
\newblock In \emph{Proceedings of the 53rd Annual Meeting of the Association
  for Computational Linguistics and the 7th International Joint Conference on
  Natural Language Processing (Volume 1: Long Papers)}, pages 344--354.

\bibitem[{Baez~Santamaria et~al.(2021)Baez~Santamaria, Baier, Kim, Krause,
  Kruijt, and Vossen}]{baez-santamaria-etal-2021-emissor}
Selene Baez~Santamaria, Thomas Baier, Taewoon Kim, Lea Krause, Jaap Kruijt, and
  Piek Vossen. 2021.
\newblock \href {https://aclanthology.org/2021.mmsr-1.6} {{EMISSOR}: A platform
  for capturing multimodal interactions as episodic memories and
  interpretations with situated scenario-based ontological references}.
\newblock In \emph{Proceedings of the 1st Workshop on Multimodal Semantic
  Representations (MMSR)}, pages 56--77, Groningen, Netherlands (Online).
  Association for Computational Linguistics.

\bibitem[{Banerjee and Lavie(2005)}]{banerjee-lavie-2005-meteor}
Satanjeev Banerjee and Alon Lavie. 2005.
\newblock \href {https://aclanthology.org/W05-0909} {{METEOR}: An automatic
  metric for {MT} evaluation with improved correlation with human judgments}.
\newblock In \emph{Proceedings of the {ACL} Workshop on Intrinsic and Extrinsic
  Evaluation Measures for Machine Translation and/or Summarization}, pages
  65--72, Ann Arbor, Michigan. Association for Computational Linguistics.

\bibitem[{Deriu et~al.(2021)Deriu, Rodrigo, Otegi, Echegoyen, Rosset, Agirre,
  and Cieliebak}]{deriu2021survey}
Jan Deriu, Alvaro Rodrigo, Arantxa Otegi, Guillermo Echegoyen, Sophie Rosset,
  Eneko Agirre, and Mark Cieliebak. 2021.
\newblock Survey on evaluation methods for dialogue systems.
\newblock \emph{Artificial Intelligence Review}, 54(1):755--810.

\bibitem[{Dinan et~al.(2018)Dinan, Roller, Shuster, Fan, Auli, and
  Weston}]{dinan2018wizard}
Emily Dinan, Stephen Roller, Kurt Shuster, Angela Fan, Michael Auli, and Jason
  Weston. 2018.
\newblock Wizard of wikipedia: Knowledge-powered conversational agents.
\newblock \emph{arXiv preprint arXiv:1811.01241}.

\bibitem[{Fitrianie et~al.(2019)Fitrianie, Bruijnes, Richards, Abdulrahman, and
  Brinkman}]{fitrianie2019we}
Siska Fitrianie, Merijn Bruijnes, Deborah Richards, Amal Abdulrahman, and
  Willem-Paul Brinkman. 2019.
\newblock What are we measuring anyway? -a literature survey of questionnaires
  used in studies reported in the intelligent virtual agent conferences.
\newblock In \emph{Proceedings of the 19th ACM International Conference on
  Intelligent Virtual Agents}, pages 159--161.

\bibitem[{Gopalakrishnan et~al.(2019)Gopalakrishnan, Hedayatnia, Chen,
  Gottardi, Kwatra, Venkatesh, Gabriel, Hakkani-T{\"u}r, and
  AI}]{gopalakrishnan2019topical}
Karthik Gopalakrishnan, Behnam Hedayatnia, Qinglang Chen, Anna Gottardi,
  Sanjeev Kwatra, Anu Venkatesh, Raefer Gabriel, Dilek Hakkani-T{\"u}r, and
  Amazon~Alexa AI. 2019.
\newblock Topical-chat: Towards knowledge-grounded open-domain conversations.
\newblock In \emph{INTERSPEECH}, pages 1891--1895.

\bibitem[{Gunasekara et~al.(2020)Gunasekara, Kim, D'Haro, Rastogi, Chen, Eric,
  Hedayatnia, Gopalakrishnan, Liu, Huang et~al.}]{gunasekara2020overview}
Chulaka Gunasekara, Seokhwan Kim, Luis~Fernando D'Haro, Abhinav Rastogi,
  Yun-Nung Chen, Mihail Eric, Behnam Hedayatnia, Karthik Gopalakrishnan, Yang
  Liu, Chao-Wei Huang, et~al. 2020.
\newblock Overview of the ninth dialog system technology challenge: Dstc9.
\newblock \emph{arXiv preprint arXiv:2011.06486}.

\bibitem[{Howcroft et~al.(2020)Howcroft, Belz, Clinciu, Gkatzia, Hasan,
  Mahamood, Mille, van Miltenburg, Santhanam, and
  Rieser}]{howcroft-etal-2020-twenty}
David~M. Howcroft, Anya Belz, Miruna-Adriana Clinciu, Dimitra Gkatzia, Sadid~A.
  Hasan, Saad Mahamood, Simon Mille, Emiel van Miltenburg, Sashank Santhanam,
  and Verena Rieser. 2020.
\newblock \href {https://aclanthology.org/2020.inlg-1.23} {Twenty years of
  confusion in human evaluation: {NLG} needs evaluation sheets and standardised
  definitions}.
\newblock In \emph{Proceedings of the 13th International Conference on Natural
  Language Generation}, pages 169--182, Dublin, Ireland. Association for
  Computational Linguistics.

\bibitem[{Li et~al.(2019)Li, Weston, and Roller}]{li2019acute}
Margaret Li, Jason Weston, and Stephen Roller. 2019.
\newblock Acute-eval: Improved dialogue evaluation with optimized questions and
  multi-turn comparisons.
\newblock \emph{arXiv preprint arXiv:1909.03087}.

\bibitem[{Lin and Och(2004)}]{lin2004looking}
Chin-Yew Lin and FJ~Och. 2004.
\newblock Looking for a few good metrics: Rouge and its evaluation.
\newblock In \emph{Ntcir workshop}.

\bibitem[{Liu et~al.(2019)Liu, Ott, Goyal, Du, Joshi, Chen, Levy, Lewis,
  Zettlemoyer, and Stoyanov}]{liu2019roberta}
Yinhan Liu, Myle Ott, Naman Goyal, Jingfei Du, Mandar Joshi, Danqi Chen, Omer
  Levy, Mike Lewis, Luke Zettlemoyer, and Veselin Stoyanov. 2019.
\newblock Roberta: A robustly optimized bert pretraining approach.
\newblock \emph{arXiv preprint arXiv:1907.11692}.

\bibitem[{Mehri and Eskenazi()}]{mehriunsupervised}
Shikib Mehri and Maxine Eskenazi.
\newblock \href {https://aclanthology.org/2020.sigdial-1.28.pdf} {Unsupervised
  evaluation of interactive dialog with dialogpt}.

\bibitem[{Mehri and Eskenazi(2020)}]{mehri-eskenazi-2020-usr}
Shikib Mehri and Maxine Eskenazi. 2020.
\newblock \href {https://doi.org/10.18653/v1/2020.acl-main.64} {{USR}: An
  unsupervised and reference free evaluation metric for dialog generation}.
\newblock In \emph{Proceedings of the 58th Annual Meeting of the Association
  for Computational Linguistics}, pages 681--707, Online. Association for
  Computational Linguistics.

\bibitem[{Mehri et~al.(2022)Mehri, Feng, Gordon, Alavi, Traum, and
  Eskenazi}]{mehri2022interactive}
Shikib Mehri, Yulan Feng, Carla Gordon, Seyed~Hossein Alavi, David Traum, and
  Maxine Eskenazi. 2022.
\newblock Interactive evaluation of dialog track at dstc9.
\newblock pages 5731‑--5738.

\bibitem[{Papineni et~al.(2002)Papineni, Roukos, Ward, and
  Zhu}]{papineni2002bleu}
Kishore Papineni, Salim Roukos, Todd Ward, and Wei-Jing Zhu. 2002.
\newblock \href {https://aclanthology.org/P02-1040.pdf} {Bleu: a method for
  automatic evaluation of machine translation}.
\newblock In \emph{Proceedings of the 40th annual meeting of the Association
  for Computational Linguistics}, pages 311--318.

\bibitem[{Pernisch et~al.(2020)Pernisch, Dell'Aglio, Serbak, and
  Bernstein}]{pernisch2020chimp}
Romana Pernisch, Daniele Dell'Aglio, Mirko Serbak, and Abraham Bernstein. 2020.
\newblock Chimp: Visualizing ontology changes and their impact in
  prot{\'e}g{\'e}.
\newblock In \emph{Fifth International Workshop on Visualization and
  Interaction for Ontologies and Linked Data}, pages 47--60. CEUR Workshop
  Proceedings.

\bibitem[{Poria et~al.(2018)Poria, Hazarika, Majumder, Naik, Cambria, and
  Mihalcea}]{poria2018meld}
Soujanya Poria, Devamanyu Hazarika, Navonil Majumder, Gautam Naik, Erik
  Cambria, and Rada Mihalcea. 2018.
\newblock Meld: A multimodal multi-party dataset for emotion recognition in
  conversations.
\newblock \emph{arXiv preprint arXiv:1810.02508}.

\bibitem[{Rashkin et~al.(2018)Rashkin, Smith, Li, and
  Boureau}]{rashkin2018towards}
Hannah Rashkin, Eric~Michael Smith, Margaret Li, and Y-Lan Boureau. 2018.
\newblock Towards empathetic open-domain conversation models: A new benchmark
  and dataset.
\newblock \emph{arXiv preprint arXiv:1811.00207}.

\bibitem[{Roller et~al.(2020)Roller, Dinan, Goyal, Ju, Williamson, Liu, Xu,
  Ott, Shuster, Smith et~al.}]{roller2020recipes}
Stephen Roller, Emily Dinan, Naman Goyal, Da~Ju, Mary Williamson, Yinhan Liu,
  Jing Xu, Myle Ott, Kurt Shuster, Eric~M Smith, et~al. 2020.
\newblock Recipes for building an open-domain chatbot.
\newblock \emph{arXiv preprint arXiv:2004.13637}.

\bibitem[{Vasiliev(2020)}]{vasiliev2020natural}
Yuli Vasiliev. 2020.
\newblock \emph{Natural Language Processing with Python and SpaCy: A Practical
  Introduction}.
\newblock No Starch Press.

\bibitem[{Vossen et~al.(2019)Vossen, Baez, Baj{\v c}eti{\'c}, Ba{\v s}i{\'c},
  and Kraaijeveld}]{vossen_leolani_2019}
Piek Vossen, Selene Baez, Lenka Baj{\v c}eti{\'c}, Suzana Ba{\v s}i{\'c}, and
  Bram Kraaijeveld. 2019.
\newblock \href {http://ceur-ws.org/Vol-2456/paper47.pdf} {Leolani: A robot
  that communicates and learns about the shared world}.
\newblock In \emph{ISWC 2019 Satellites}, CEUR Workshop Proceedings, pages
  181--184. CEUR-WS.
\newblock 2019 ISWC Satellite Tracks (Posters and Demonstrations, Industry, and
  Outrageous Ideas), ISWC 2019-Satellites ; Conference date: 26-10-2019 Through
  30-10-2019.

\bibitem[{Vossen et~al.(2018)Vossen, Baez, Baj\v{c}eti\'{c}, and
  Kraaijeveld}]{vossen2018leolani}
Piek Vossen, Selene Baez, Lenka Baj\v{c}eti\'{c}, and Bram Kraaijeveld. 2018.
\newblock Leolani: a reference machine with a theory of mind for social
  communication.
\newblock In \emph{International conference on text, speech, and dialogue},
  pages 15--25. Springer.

\bibitem[{Weizenbaum(1966)}]{weizenbaum1966eliza}
Joseph Weizenbaum. 1966.
\newblock Eliza—a computer program for the study of natural language
  communication between man and machine.
\newblock \emph{Communications of the ACM}, 9(1):36--45.

\bibitem[{Zhang* et~al.(2020)Zhang*, Kishore*, Wu*, Weinberger, and
  Artzi}]{bert-score}
Tianyi Zhang*, Varsha Kishore*, Felix Wu*, Kilian~Q. Weinberger, and Yoav
  Artzi. 2020.
\newblock \href {https://openreview.net/forum?id=SkeHuCVFDr} {Bertscore:
  Evaluating text generation with bert}.
\newblock In \emph{International Conference on Learning Representations}.

\bibitem[{Zhang et~al.(2019)Zhang, Sun, Galley, Chen, Brockett, Gao, Gao, Liu,
  and Dolan}]{zhang2019dialogpt}
Yizhe Zhang, Siqi Sun, Michel Galley, Yen-Chun Chen, Chris Brockett, Xiang Gao,
  Jianfeng Gao, Jingjing Liu, and Bill Dolan. 2019.
\newblock Dialogpt: Large-scale generative pre-training for conversational
  response generation.
\newblock \emph{arXiv preprint arXiv:1911.00536}.

\end{thebibliography}
\bibliographystyle{acl_natbib}

\newpage
\appendix

\section{Appendix}

\subsection{Leolani, the agent}
\label{sec:appendix-agents}

The Leolani agent creates an eKG on the fly and formulates a response to a prompt based on the changes in the graph. At the start, an empty eKG is initialized with the Nice-to-Meet-You (N2MU) ontology~\cite{vossen_leolani_2019} that defines basic things in the world such as people, locations, and some social properties, but without any knowledge about individual people. Triple extraction during the conversation adds such knowledge.

In addition to passive processing of human prompts, our agent Leolani reflects on the changing eKG during the conversation. After each change, a series of semantically meaningful SPARQL queries is applied to the eKG to detect specific graph properties that are deemed relevant to respond to. These handmade query patterns detect phenomena like conflicts, uncertainty, analogies, generalizations, knowledge gaps, trust in knowledge sources, the novelty of factoids, and the relevance of factoids to interlocutors. We consider these queries as "\textit{thoughts}" that reflect on the current state of the eKG, given the latest incoming information. Each type of \textit{thought} is associated to a broad higher level goal, which we call "\textit{drives}", for example, to improve the correctness or completeness of the agent's current knowledge, or to improve the engagement or social connection to the interlocutor. These \textit{drives} result in speech acts such as statements or questions that the agent directs to the interlocutor, aiming of obtaining a relevant response. In the next short dialog, the agent Leolani learned a new concept "reading" and tries to generalise this to all people by asking a question to the human. The next statement by the human triggers the agent to ask another follow-up question:

\begin{itemize}

\itemsep0em 
\item H: I like reading 
\item A: I never heard about reading before! can you tell me, is reading something people do?
\item H: Yes, people read books
\item A: I did not know that, what else do people read?

\end{itemize}

\newpage
\subsection{Context-Free-Grammar Triple Extractor}
\label{sec:appendix-extractors }

Triple extraction is performed by a dependency parser with a Context-Free Grammar (CFG) consisting of 12 rules and a lexicon with mainly closed-class words. It is adapted to resolve deictic references to the addressee or other observed objects in a situation. As the conversation is open and can be about any topic, part-speech-tagging and named entity-recognition is used to detect patterns of open-class words as subjects, predicates, and objects for triples. 


\newpage
\subsection{Sample student-Leolani conversations}
\label{sec:appendix-sample}

\begin{lstlisting}[numbers=none,language=MyDialogue, caption=Extract from conversation with Student 1]
Leolani: "What's up? What is your name? Stranger?"

Student1: "Student1"

Leolani: "So your name is Student1?"

Student1: "yes"

Leolani: "Would you like to chat? I'll do my best to keep up"

Student1: "I like chatting"

Leolani: "I would like to know. What types of Instance like chatting do person or Instance usually like"

Student1: "Person likes convos"

Leolani: "I am curious. What types of noun or Instance like convos do tops or Instance usually like"

Student1: "Convos are not people"

Leolani: "Let me ask you something. Has convos work at institution?"

Student1: "I work at institution"

Leolani: "Let me ask you something. Has thomas experience touch?"

Student1: "I have experience touch"

Leolani: "I am curious. Has thomas own object?"

Student1: "I own object"

Leolani: "Interesting! I am excited to get to know about you!"

Student1: "What do you want to know?"

Leolani: "I know agent usually want to verb.cognition, but I do not know this case"

Student1: "I like sushi"

Leolani: "Exciting news! I did not know anything that Student1 like"

Student1: "I also like cats"

Leolani: "If you don't mind me asking. What types of animal or Instance like cats do person or Instance usually like"
\end{lstlisting}

\newpage
\subsection{Analysis of student-Leolani conversations}
\label{sec:appendix-triple-details}

In total, 81,631 triples were generated from the student conversations. Note that not all triples count as factual knowledge. 23 predicates were extracted of which "know" and "be" are most frequent (Table~\ref{tab:pred}). The "know" predicates are mostly derived from introducing oneself to the agent. The predicates further show that the communication is open, although the semantics is not further defined beyond the predicate itself, e.g. there is no relation between "smell" and "can-smell".  The triples further contained 31 unique subjects and 78 unique objects; most occurred only once in the communication (Table~\ref{tab:sub-obj}). 

\begin{table}[!h]
    \centering
    \scriptsize{
        \caption{Predicate labels and their frequency in student conversations.}
    \label{tab:pred}
    \begin{tabular}{|l|l|l|l|l|l|}
    \hline
        know & 43 & can-fly & 2 & favourite-animal-is & 1\\ \hline
        be & 39 & like-to & 2 & favourite-cat-is & 1 \\ \hline
        like & 22 & work-at & 2 & fly & 1 \\ \hline
        live-in & 9 & be-in & 1 & hair-color-is & 1 \\ \hline
        sense & 7 & be-to & 1 & own & 1 \\ \hline
        have & 6 & can-learn & 1 & smell & 1 \\ \hline
        be-from & 3 & can-smell & 1 & wear & 1 \\ \hline
        love & 3 & could-help & 1 && \\ \hline
    \end{tabular}}
\end{table}

\begin{table}[!h]
    \centering
    \scriptsize{
        \caption{Entity labels and their frequency in student conversations.}
    \label{tab:sub-obj}
    \begin{tabular}{|p{1.3cm}|p{.1cm}|p{1.6cm}|p{.1cm}|p{1.6cm}|p{.1cm}|}
    \hline
        Leolani & 23 & a-dog & 1 & go & 1 \\ \hline
        student1 & 11 & a-girl & 1 & great-that-i & 1 \\ \hline
        student2 & 10 & a-man & 1 & his-uncle & 1 \\ \hline
        student3 & 8 & a-shame & 1 & institution & 1  \\ \hline
        student4 & 8 & a-student & 1 & student10 & 1  \\ \hline
        student5 & 8 & a-wise-man & 1 & know & 1  \\ \hline
        student6 & 7 & a-woman & 1 & lasagne & 1  \\ \hline
        a-flamingo & 6 & airplanes & 1 & phd2-student5 & 1 \\ \hline
        student7 & 6 & amstelveen & 1 &  phd2-name & 1 \\ \hline
        student8 & 5 & an-animal & 1 & Leolani-new-things & 1 \\ \hline
        amsterdam & 4 & an-aunt & 1 & my-daughter & 1  \\ \hline
        student9 & 4 & an-emotion & 1 & my-parents & 1 \\ \hline
        my & 3 & an-uncle & 1 & student5-friend & 1 \\ \hline
        orange & 3 & brown & 1 & now & 1  \\ \hline
        reading & 3 & bulgaria & 1 & object & 1 \\ \hline
        alkmaar & 2 & business-class & 1 & other-things & 1  \\ \hline
        cats & 2 & candy & 1 & parents & 1  \\ \hline
        chatting & 2 & city & 1 & people & 1 \\ \hline
        convos & 2 & cook & 1 & person & 1  \\ \hline
        garfield & 2 & cook-by-myself & 1 & pink & 1  \\ \hline
        japanese-food & 2 & deloitte & 1 & rotterdam & 1  \\ \hline
        love & 2 & dogs & 1 & shrimp & 1 \\ \hline
        None & 2 & every-agent & 1 & student & 1 \\ \hline
        phd1 & 2 & experience-touch & 1 & sushi & 1 \\ \hline
        10-fingers & 1 & favorite-of-student6 & 1 & tapas & 1 \\ \hline
        a-bird & 1 & flamingo & 1 & the-south-of-holland & 1 \\ \hline
        a-color & 1 & food & 1 & two-hands & 1 \\ \hline
        a-company & 1 & front-camera & 1 & what & 1 \\ \hline
        a-country & 1 & garfield-favourite-food & 1 & yes-candy & 1 \\ \hline
        a-daughter & 1 & glasses & 1 & ~ & ~ \\ \hline
    \end{tabular}}
\end{table}

\newpage
\subsection{Full list of metrics tested}
The full list of 62 metrics used, sorted by group:

\begin{itemize}
\small{
\itemsep0em 
\item GROUP A
\begin{itemize}
    \item Total nodes
    \item Total edges
    \item Average degree
    \item Average degree centrality
    \item Average closeness
    \item Average betweenness
    \item Average degree connectivity
    \item Average assortativity
    \item Average node connectivity
    \item Number of components
    \item Number of strong components
    \item Shortest path
    \item Centrality entropy
    \item Closeness entropy
    \item Sparseness
\end{itemize}
\item GROUP B
\begin{itemize}
    \item Total classes
    \item Total properties
    \item Total instances
    \item Total object properties
    \item Total data properties
    \item Total equivalent class properties
    \item Total subclass properties
    \item Total entities
    \item Total inverse entities
    \item Ratio of inverse relations
    \item Property class ratio
    \item Average population
    \item Class property ratio
    \item Attribute richness
    \item Inheritance richness
    \item Relationship richness
    \item Object properties ratio
    \item Datatype properties ratio
    \item Total concept assertions
    \item Total role assertions
    \item Total general concept inclusions
    \item Total domain axioms
    \item Total range axioms
    \item Total role inclusions
    \item Total axioms
    \item Total aBox axioms
    \item Total tBox axioms
\end{itemize}
\item GROUP C
\begin{itemize}
    \item Total triples
    \item Total world instances
    \item Total claims
    \item Total perspectives
    \item Total mentions
    \item Total conflicts
    \item Total sources
    \item Total interactions
    \item Total utterances
    \item Ratio claim to triples
    \item Ratio perspectives to triples
    \item Ratio conflict to triples
    \item Ratio perspectives to claims
    \item Ratio mentions to claims
    \item Ratio conflicts to claims
    \item Average perspectives per claim
    \item Average mentions per claim
    \item Average turns per interaction
    \item Average claims per source
    \item Average perspectives per source
\end{itemize}
}
\end{itemize}

\newpage
\subsection{Correlation matrix}
Hereby we show the full correlation matrix between our proposed metrics and the human and automatic evaluations. Metrics related to conflicts are not informative since the short conversations did not produce conflicting information. Metrics related to perspectives are hindered because the simple triple extractor is limited in the range of perspectives it can extract.

\begin{figure*}[!h]
    \centering
    \includegraphics[trim={0 0 0 0},clip,width=0.85\textwidth]{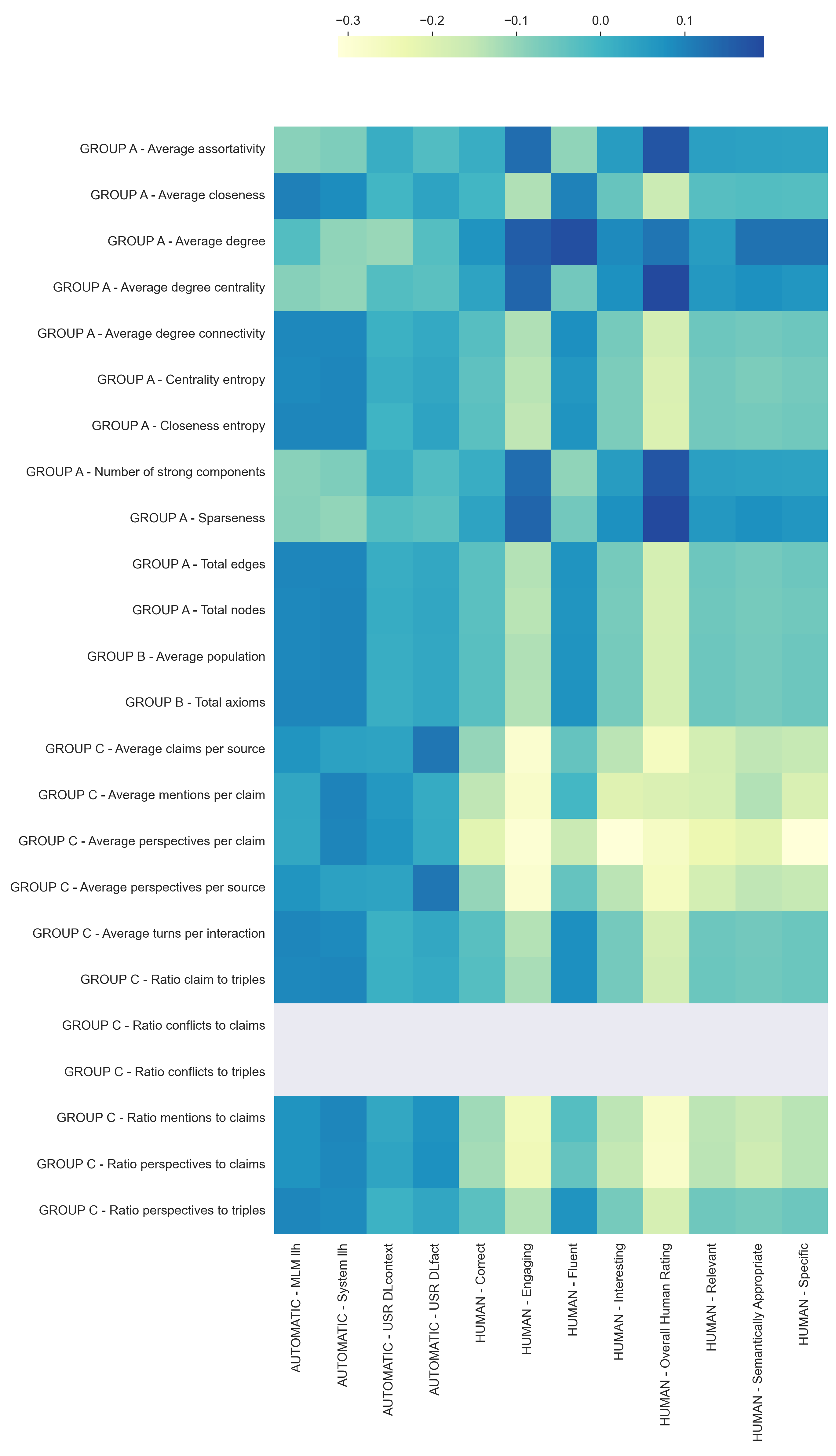}
    \caption{Correlation matrix for evaluation metrics.}
    \label{fig:correl}
\end{figure*}

\end{document}